\documentclass{article}
\usepackage{ijcai20}

\usepackage{times}

\usepackage{soul}
\usepackage{url}
\usepackage[hidelinks]{hyperref}
\usepackage[utf8]{inputenc}
\usepackage[small]{caption}
\usepackage{graphicx}
\usepackage{amsmath}
\usepackage{booktabs}
\usepackage{algorithm}
\usepackage{algorithmic}
\usepackage{url,ifthen}
\usepackage{amssymb}
\usepackage{fancybox}
\usepackage{tikz}
\usepackage{enumitem}

\usepackage{psgo}

\title{Monte Carlo Game Solver}

\author{
    Tristan Cazenave
    \affiliations
    LAMSADE, Université Paris-Dauphine, PSL, CNRS, France
    \emails
    Tristan.Cazenave@dauphine.psl.eu
}

\begin{document}

\maketitle

\begin{abstract}
We present a general algorithm to order moves so as to speedup exact game
solvers. It uses online learning of playout policies and Monte Carlo
Tree Search. The learned policy and the information in the Monte Carlo 
tree are used to order moves in game solvers. They improve greatly the 
solving time for multiple games.
\end{abstract}

\section{Introduction}

Monte Carlo Tree Search (MCTS) associated to Deep Reinforcement learning has superhuman results in the most difficult complete information games (Go, Chess and Shogi) \cite{silver2018general}. However little has been done to use this kind of algorithms to exactly solve games. We propose to use MCTS associated to reinforcement learning of policies so as to speedup the resolution of various games.

The paper is organized as follows: the second section deals with
related work on games. The third section details the move ordering
algorithms for various games. The fourth section gives experimental
results for these games.

\section{Previous Work}

In this section we review the different algorithms that have been used to solve games. We then focus on the $\alpha\beta$ solver. As we improve $\alpha\beta$ with MCTS we show the difference to MCTS Solver. We also expose Depth First Proof Number Search as it has solved multiple games. We finish with a description of online policy learning as the resulting policy is used for our move ordering.

\subsection{Solving Games}

Iterative Deepening Alpha-Beta associated to a heuristic evaluation
function and a transposition table is the standard algorithm for
playing games such as Chess and Checkers. Iterative Deepening
Alpha-Beta has also been used to solve games such as small board Go
\cite{van2009solving}, Renju \cite{WagnerRenju01}, Lines of Action
\cite{SakutaLOA03}, Amazons endgames \cite{kloetzer2008comparative}.
Other researchers have instead used a Depth-first Alpha-Beta without
Iterative Deepening and with domain specific algorithms to solve
Domineering \cite{Uiterwijk2016} and Atarigo \cite{Boissac2006}. The
advantage of Iterative Deepening associated to a transposition table
for solving games is that it finds the shortest win and that it reuses
the information of previous iterations for ordering the moves, thus
maximizing the cuts. The heuristics usually used to order the move in
combination with Iterative Deepening are: trying first the
transposition table move, then trying killer moves and then sorting
the remaining moves according to the History
Heuristic \cite{Schaeffer1989history}. The advantage of not using
Iterative Deepening is that the iterations before the last one are not
performed, saving memory and time, however if bad choices on move
ordering happen, the search can waste time in useless parts of the
search tree and can also find move sequences longer than necessary.

There are various competing algorithms for solving games
\cite{Herik2002Games}. The most simple are Alpha-Beta and Iterative
Deepening Alpha-Beta. Other algorithms memorize the search tree in
memory and expand it with a best first strategy: Proof Number Search
\cite{Allis1994PN}, PN$^2$ \cite{Breuker1998}, Depth-first Proof
Number Search (Df-pn) \cite{Nagai2002}, Monte Carlo Tree Search Solver
\cite{winands2008monte,CazenaveS2010} and Product Propagation
\cite{SaffidineCazenave2013CG}.

Games solved with a best first search algorithm include Go-Moku with
Proof Number search and Threat Space Search \cite{Allis1996}, Checkers
using various algorithms \cite{Schaeffer2007checkers}, Fanorona with
PN$^2$
\cite{Schadd2008best}, $6 \times 6$ Lines of Action with PN$^2$
\cite{Winands2008LOA}, $6 \times 5$ Breakthrough with parallel PN$^2$
\cite{Saffidine2011Solving}, and $9 \times 9$ Hex with parallel Df-pn
\cite{Pawlewicz2013}.

Other games such as Awari were solved using retrograde analysis
\cite{Romein2003}. Note that retrograde analysis was combined with
search to solve Checkers and Fanorona.

\subsection{$\alpha\beta$ Solver}

Iterative Deepening $\alpha\beta$ has long been the best algorithm for multiple games. Most of the Chess engines still use it even if the current best algorithm is currently MCTS \cite{silver2018general}.

Depth first $\alpha\beta$ is more simple but it can be better than Iterative Deepening $\alpha\beta$ for solving games since it does not have to explore a large tree before searching the next depth. More over in the case of games with only two outcomes the results are always either Won or Lost and enable immediate cuts when Iterative Deepening $\alpha\beta$ has to deal with unknown values when it reaches depth zero and the state is not terminal.

One interesting property of $\alpha\beta$ is that selection sort becomes an interesting sorting algorithm. It is often useful to only try the best move or a few good moves before reaching a cut. Therefore it is not necessary to sort all the moves at first. Selecting move by move as in selection sorting can be more effective.

\begin{algorithm}
\begin{algorithmic}[1]
\STATE{Function $\alpha\beta$ ($s$,$depth$,$\alpha$,$\beta$)}
\begin{ALC@g}
\IF{isTerminal ($s$) \OR $depth = 0$}
\RETURN Evaluation ($s$)
\ENDIF
\IF{$s$ has an entry $t$ in the transposition table}
\IF{the result of $t$ is exact}
\RETURN $t.res$
\ENDIF
\STATE{put $t.move$ as the first legal move}
\ENDIF
\FOR{$move$ in legal moves for $s$}
\STATE{$s_1$ = play ($s$, $move$)}
\STATE{$eval$ = -$\alpha\beta$($s_1$,$depth-1$,-$\beta$,-$\alpha$)}
\IF{$eval > \alpha$}
\STATE{$\alpha = eval$}
\ENDIF
\IF{$\alpha \geq \beta$}
\STATE{update the transposition table}
\RETURN $\beta$
\ENDIF
\ENDFOR
\STATE{update the transposition table}
\RETURN $\alpha$
\end{ALC@g}
\end{algorithmic}
\caption{\label{AlphaBeta}The $\alpha\beta$ algorithm for solving games.}
\end{algorithm}

\subsection{MCTS Solver}

MCTS has already been used as a game solver \cite{winands2008monte}. The principle is to mark as solved the subtrees that have an exact result. As the method uses playouts it has to go through the tree at each playout and it revisits many times the same states doing costly calculations to choose the move to try according to the bandit. Moreover in order to solve a game a large game tree has to be kept in memory.

The work on MCTS Solver has later been extended to games with multiple outcomes \cite{CazenaveS2010}.

\subsection{Depth First Proof Number Search}

Proof Number Search is a best first algorithm that keeps the search tree in memory so as to expand the most informative leaf \cite{Allis1994PN}. In order to solve the memory problem of Proof Number Search, the PN$^2$ algorithm has been used \cite{Breuker1998}. PN$^2$ uses a secondary Proof Number Search at each leaf of the main Proof Number Search tree, thus enabling the square of the total number of nodes of the main search tree to be explored. More recent developments of Proof Number Search focus on Depth-First Proof Number search (DFPN) \cite{Nagai2002}. The principle is to use a transposition table and recursive depth first search to efficiently search the game tree and solve the memory problems of Proof Number Search. DFPN can be parallelized to improve the solving time \cite{hoki2013parallel}. It has been improved for the game of Hex using a trained neural network  \cite{gao2017focused}. It can be applied to many problems, including recently Chemical Synthesis Planning \cite{kishimoto2019depth}.

\subsection{Online Policy Learning}

Playout Policy Adaptation with Move Features (PPAF) has been applied to many games \cite{CazenavePPAF16}.

An important detail of the playout algorithm is the code function. In PPAF
the same move can have different codes that depend on the presence of
features associated to the move. For example in Breakthrough the code also takes into account whether the arriving square is empty or contains an opponent pawn.

The principle of the learning algorithm is to add 1.0 to the weight of
the moves played by the winner of the playout. It also decreases the
weights of the moves not played by the winner of the playout by a
value proportional to the exponential of the weight. This algorithm is given in algorithm \ref{ADAPT}.

\begin{algorithm}
\begin{algorithmic}[1]
\STATE{Function adapt ($winner$, $board$, $player$, $playout$, $policy$)}
\begin{ALC@g}
\STATE{$polp \leftarrow$ $policy$}
\FOR{$move$ in $playout$}
\IF{$winner$ = $player$}
\STATE{$polp$ [code($move$)] $\leftarrow$ $polp$ [code($move$)] + $\alpha$}
\STATE{$z$ $\leftarrow$ 0.0}
\FOR{$m$ in possible moves on $board$}
\STATE{$z$ $\leftarrow$ $z$ + exp ($policy$ [code($m$)])}
\ENDFOR
\FOR{$m$ in possible moves on $board$}
\STATE{$polp$ [code($m$)] $\leftarrow$ $polp$ [code($m$)] - $\alpha * \frac{exp (policy [code(m)])}{z}$}
\ENDFOR
\ENDIF
\STATE{play ($board$, $move$)}
\STATE{$player$ $\leftarrow$ opponent ($player$)}
\ENDFOR
\STATE{$policy$ $\leftarrow$ $polp$}
\end{ALC@g}
\end{algorithmic}
\caption{\label{ADAPT}The PPAF adapt algorithm}
\end{algorithm}

\section{Move Ordering}

We describe the general tools used for move ordering then their adaptation to different games.

\subsection{Outline}

In order to collect useful information to order moves we use a combination of the GRAVE algorithm \cite{GRAVE} and of the PPAF algorithm. Once the Monte Carlo search is finished we use the transposition table  of the Monte Carlo search to order the moves, putting first the most simulated ones. When outside the transposition table we use the learned weights to order the moves. The algorithm used to score the moves so as to order them is given in algorithm \ref{sorting}.

\begin{algorithm}
\begin{algorithmic}[1]
\STATE{Function orderMC ($board$, $code$)}
\begin{ALC@g}
\STATE{$ppaf \leftarrow policy [code]$}
\IF{$board$ has an entry $t$ in the MCTS TT}
\IF{$t.nbPlayouts > 100$}
\FOR{$move$ in legal moves for $board$}
\IF{$t.nbPlayouts [move] > 0$}
\IF{$code (move) = code$}
\STATE{$ppaf \leftarrow t.nbPlayouts [move]$}
\ENDIF
\ENDIF
\ENDFOR
\ENDIF
\ENDIF
\RETURN $1000000000 - 1000 \times ppaf$
\end{ALC@g}
\end{algorithmic}
\caption{\label{sorting}The Monte Carlo Move Ordering function}
\end{algorithm}

\subsection{Atarigo}

Atarigo is a simplification of the game of Go. The first player to capture has won. It is a game often used to teach Go to beginners. Still it is an interesting games and tactics can be hard to master.

The algorithm for move ordering is given in algorithm \ref{atarigo}. It always puts first a capture move since it wins the game. If no such move exist it always plays a move that saves a one liberty string since it is a forced move to avoid losing. Then it favors moves on liberties of opponent strings that have few liberties provided the move has itself sufficient liberties. If none of these are available it returns the evaluation by the Monte Carlo ordering function.

The code associated to a move is built after its four neighboring intersections.

\begin{algorithm}
\begin{algorithmic}[1]
\STATE{Function order ($board$, $move$)}
\begin{ALC@g}
\STATE{$minOrder \leftarrow 361$}
\FOR{$i$ in adjacents to $move$}
\IF{$i$ is an opponent stone}
\STATE{$n \leftarrow$ number of liberties of $i$}
\IF{$n = 1$}
\RETURN 0
\ENDIF
\STATE{$nb \leftarrow n - 4 \times nbEmptyAdjacent (move)$}
\IF{$nb < minOrder$}
\STATE{$minOrder \leftarrow nb$}
\ENDIF
\ENDIF
\ENDFOR
\IF{$move$ escapes an atari}
\RETURN 1
\ENDIF
\IF{$minOrder = 361$}
\IF{MonteCarloMoveOrdering}
\RETURN $orderMC (board, code(move))$
\ENDIF
\RETURN $20-nbEmptyAdjacent (move)$
\ENDIF
\RETURN $minOrder$
\end{ALC@g}
\end{algorithmic}
\caption{\label{atarigo}The function to order moves at Atarigo}
\end{algorithm}

\subsection{Nogo}

Nogo is the misere version of Atarigo \cite{ChouTY2011}. It was
introduced at the 2011 Combinatorial Game Theory Workshop in
Banff. The first player to capture has lost. It is usually played on
small boards. In Banff there was a tournament for programs and Bob
Hearn won the tournament using the Fuego framework
\cite{enzenberger2010fuego} and Monte-Carlo Tree Search.

We did not find simple heuristics to order moves at Nogo. So the
standard algorithm uses no heuristic and the MC algorithms sort moves
according to algorithm \ref{sorting}.

\subsection{Go}

Go was solved for rectangular boards up to size $7 \times 4$ by the MIGOS program \cite{van2009solving}. The algorithm used was an iterative deepening $\alpha\beta$ with transposition table.

\subsection{Breakthrough and Knightthrough}

Breakthrough is an abstract strategy board game invented by Dan Troyka in 2000. It won the 2001 8x8 Game Design Competition and it is played on Little Golem. The game starts with two rows of pawns on each side of the board. Pawns can capture diagonally and go forward either vertically or diagonally. The first player to reach the opposite row has won. Breakthrough has been solved up to size $6 \times 5$ using Job Level Proof Number Search \cite{Saffidine2011Solving}. The code for a move at Breakthrough contains the starting square, the arrival square and whether it is empty or contains an enemy pawn. The ordering gives priority to winning moves, then to moves to prevent a loss, then Monte Carlo Move Ordering.

Misere Breakthrough is the misere version of Breakthrough, the games is lost if a pawn reaches the opposite side. It is also a difficult game and its is more difficult for MCTS algorithms \cite{CazenavePPAF16}. The code for a move is the same as for Breakthrough and the ordering is Monte Carlo Move Ordering.

Knightthrough emerged as a game invented for the General Game Playing competitions. Pawns are replaced with knights. Misere Knightthrough is the misere version of the game where the goal is to lose. Codes for moves and move ordering are similar to Breakthrough.

\begin{algorithm}
\begin{algorithmic}[1]
\STATE{Function order ($board$, $move$)}
\begin{ALC@g}
\IF{$move$ is a winning move}
\RETURN 0
\ENDIF
\IF{$move$ captures an opponent piece}
\IF{capture in the first 3 lines}
\RETURN 1
\ENDIF
\ENDIF
\IF{destination in the last 3 lines}
\IF{support(destination) $>$ attack(destination)}
\RETURN 2
\ENDIF
\ENDIF
\IF{MonteCarloMoveOrdering}
\RETURN $orderMC (board, code(move))$
\ENDIF
\RETURN 100
\end{ALC@g}
\end{algorithmic}
\caption{\label{knightthrough}The function to order moves at Knightthrough}
\end{algorithm}

\subsection{Domineering}

Domineering is played on a chess board and two players alternate putting dominoes on the board. The first player puts the dominoes vertically, the second player puts them horizontally. The first player who cannot play loses. In Misere Domineering the first player who cannot play wins.

\section{Experimental results}

The iterative deepening $\alpha\beta$ with a transposition table (ID
$\alpha\beta$ TT) is called with a null window since it saves much
time compared to calling it with a full window. Other algorithms are
called with the full window since it they only deal with terminal
states values and that the games we solve are either Won or Lost.

A transposition table containing 1 048 575 entries is used for all
games. An entry in the transposition table is always replaced by a new
one.

An algorithm name finishing with MC denotes the use of Monte Carlo
Move Ordering. The times given for MC algorithms include the time for
the initial MCTS that learns a policy. The original Proof Number
Search algorithm is not included in the experiments since it fails due
to being short of memory for complex games. The $PN^{2}$ algorithm
solves this memory problem and is included in the experiments. 
The algorithms that do not use MC still do some move ordering but
without Monte Carlo.


\begin{table}
  \centering
  \caption{Different algorithms for solving Atarigo.}
  \label{tableAtarigo}
  \begin{tabular}{lrrrrrrrrrr}
                       &                    &                 \\ 
Size                   &       $5 \times 5$ &                 \\ 
Result                 &                Won &                 \\ 
                       &              Moves &            Time \\ 
$PN^{2}$                &     14 784 088 742 &     37 901.56 s.\\ 
ID $\alpha\beta$ TT    & $>$ 35 540 000 000 & $>$ 86 400.00 s.\\ 
$\alpha\beta$ TT       & $>$ 37 660 000 000 & $>$ 86 400.00 s.\\ 
ID $\alpha\beta$ TT MC &         62 800 334 &        126.84 s.\\ 
$\alpha\beta$ TT MC    &      \bf 3 956 049 &     \bf 12.79 s.\\ 
                       &                    &                 \\ 
Size                   &       $6 \times 5$ &                 \\ 
Result                 &                Won &                 \\ 
                       &              Moves &            Time \\ 
$PN^{2}$               & $>$ 33 150 000 000 & $>$ 86 400.00 s.\\ 
ID $\alpha\beta$ TT    & $>$ 37 190 000 000 & $>$ 86 400.00 s.\\ 
$\alpha\beta$ TT       & $>$  7 090 000 000 & $>$ 44 505.91 s.\\ 
ID $\alpha\beta$ TT MC &     12 713 931 627 &     27 298.35 s.\\ 
$\alpha\beta$ TT MC    &    \bf 329 780 434 &    \bf 787.17 s.\\ 
  \end{tabular}
\end{table}

Table \ref{tableAtarigo} gives the results for Atarigo. For Atarigo
$5 \times 5$ $\alpha\beta$ TT MC is the best algorithm and is much
better than $\alpha\beta$ TT. For Atarigo $6 \times 5$ the best
algorithm is again $\alpha\beta$ TT MC which is much better than all
other algorithms.

\begin{table}
  \centering
  \caption{Different algorithms for solving Nogo.}
  \label{tableNogo}
  \begin{tabular}{lrrrrrrrrrr}
                       &                     &                 \\ 
Size                   &        $7 \times 3$ &                 \\ 
Result                 &                 Won &                 \\ 
                       &               Moves &            Time \\ 
$PN^{2}$                &  $>$ 80 390 000 000 & $>$ 86 400.00 s.\\ 
ID $\alpha\beta$ TT    &      10 921 978 839 &     12 261.64 s.\\ 
$\alpha\beta$ TT       &       3 742 927 598 &      4 412.21 s.\\ 
ID $\alpha\beta$ TT MC &       1 927 635 856 &      2 648.91 s.\\ 
$\alpha\beta$ TT MC    &      \bf 35 178 886 &     \bf 49.72 s.\\ 
                       &                     &                 \\ 
Size                   &        $5 \times 4$ &                 \\ 
Result                 &                 Won &                 \\ 
                       &               Moves &            Time \\ 
$PN^{2}$                & $>$ 101 140 000 000 & $>$ 86 400.00 s.\\ 
ID $\alpha\beta$ TT    &       1 394 182 870 &      1 573.72 s.\\ 
$\alpha\beta$ TT       &       1 446 922 704 &      1 675.64 s.\\ 
ID $\alpha\beta$ TT MC &          73 387 083 &        134.26 s.\\ 
$\alpha\beta$ TT MC    &      \bf 33 850 535 &     \bf 74.77 s.\\ 
  \end{tabular}
\end{table}

Table \ref{tableNogo} gives the results for Nogo. Nogo $7 \times 3$ is
solved in 49.72 seconds by $\alpha\beta$ TT MC with 100 000
playouts. This is 88 times faster than $\alpha\beta$ TT the best
algorithm not using MC.

Nogo $5 \times 4$ is solved best by $\alpha\beta$ TT MC with 1 000 000
playouts before the $\alpha\beta$ search. It is 21 times faster than ID
$\alpha\beta$ TT the best algorithm not using MC.

$\alpha\beta$ TT MC with 10 000 000 playouts solves Nogo $5 \times 5$
in 61 430.88 seconds and 46 092 056 485 moves. This is the first time
Nogo $5 \times 5$ is solved. The solution is:

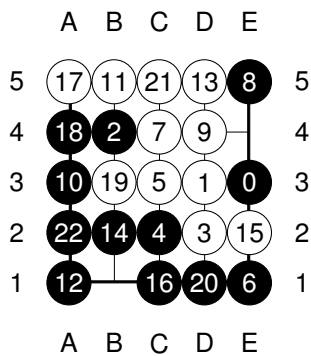
\begin{center}
\begin{psgoboard}[5]
\move{e}{3}
\move{d}{3}
\move{b}{4}
\move{d}{2}
\move{c}{2}
\move{c}{3}
\move{e}{1}
\move{c}{4}
\move{e}{5}
\move{d}{4}
\move{a}{3}
\move{b}{5}
\move{a}{1}
\move{d}{5}
\move{b}{2}
\move{e}{2}
\move{c}{1}
\move{a}{5}
\move{a}{4}
\move{b}{3}
\move{d}{1}
\move{c}{5}
\move{a}{2}
\end{psgoboard}
\end{center}

As it is the first time results about solving Nogo are given we 
recapitulate in table \ref{tableSolvedNogo} the winner for various 
sizes. A one means a first player win and a two a second player win.

\begin {table}
\begin{center}
\begin{tabular}{|c|rrrrrrrrrr|}
\hline
\rule{0pt}{12pt}
 & 1 & 2 & 3 & 4 & 5 & 6 & 7 & 8 & 9 & 10
\\
\hline
 1 & 2 & 1 & 1 & 2 & 1 & 1 & 1 & 1 & 1 & 1 \\
 2 & 1 & 1 & 2 & 2 & 1 & 1 & 1 & 1 & 2 & 2 \\
 3 & 1 & 2 & 1 & 2 & 1 & 1 & 1 & 1 &   &   \\
 4 & 2 & 2 & 2 & 2 & 1 & 1 &   &   &   &   \\
 5 & 1 & 1 & 1 & 1 & 1 &   &   &   &   &   \\
 6 & 1 & 1 & 1 & 1 &   &   &   &   &   &   \\
 7 & 1 & 1 & 1 &   &   &   &   &   &   &   \\
 8 & 1 & 1 & 1 &   &   &   &   &   &   &   \\
 9 & 1 & 2 &   &   &   &   &   &   &   &   \\
10 & 1 & 2 &   &   &   &   &   &   &   &   \\
\hline
\end{tabular}
{\caption{Winner for Nogo boards of various sizes}\label{tableSolvedNogo}}
\end{center}
\end{table}

\begin{table}
  \centering
  \caption{Different algorithms for solving Go.}
  \label{tableGo}
  \begin{tabular}{lrrrrrrrrrr}
                       &                    &                 \\ 
Size                   &       $3 \times 3$ &                 \\ 
Result                 &                Won &                 \\ 
                       &              Moves &            Time \\ 
$PN^{2}$               &            246 394 &   3.72 s.\\ 
ID $\alpha\beta$ TT    &            840 707 &  11.34 s.\\ 
$\alpha\beta$ TT       &            420 265 &  11.50 s.\\ 
ID $\alpha\beta$ TT MC &            375 414 &   5.62 s.\\ 
$\alpha\beta$ TT MC    &         \bf 6 104  & \bf 0.16 s.\\ 
                       &                    &                 \\ 
Size                   &       $4 \times 3$ &                 \\ 
Result                 &                Won &                 \\ 
                       &              Moves &            Time \\ 
$PN^{2}$               &  43 202 038 &   619.98 s.\\ 
ID $\alpha\beta$ TT    &  39 590 950 &   515.71 s.\\ 
$\alpha\beta$ TT       & 107 815 563 & 1 977.86 s.\\ 
ID $\alpha\beta$ TT MC &  22 382 730 &   348.08 s.\\ 
$\alpha\beta$ TT MC    & \bf 4 296 893 & \bf 96.63 s.\\ 
  \end{tabular}
\end{table}

Table \ref{tableGo} gives the results for Go. Playouts and depth first $\alpha\beta$ can last a very long time in Go since stones are captured and if random play occurs the goban can become almost empty again a number of times before the superko rules forbids states. In order to avoid very long and useless games an artificial limit on the number of moves allowed in a game was set to twice the size of the board. This is not entirely satisfactory since one can imagine weird cases where the limit is not enough. The problem does not appear in the other games we have solved since they converge to a terminal state before a fixed number of moves. The trick we use to address the problem is to send back an evaluation of zero if the search reaches the limit. When searching for a win with a null window this is equivalent to a loss and when searching for a loss it is equivalent to a win. Therefore if the search finds a win it does not rely on the problematic states. The $3 \times 3$ board was solved with a komi of 8.5, the $4 \times 3$ board was solved with a komi of 3.5.

\begin{table}
  \centering
  \caption{Different algorithms for solving Breakthrough.}
  \label{tableBreakthrough}
  \begin{tabular}{lrrrrrrrrrr}

                       &                    &                 \\ 
Size                   &       $5 \times 5$ &                 \\ 
Result                 &               Lost &                 \\ 
                       &              Moves &            Time \\ 
$PN^{2}$                & $>$ 38 780 000 000 & $>$ 86 400.00 s.\\ 
ID $\alpha\beta$ TT    &     13 083 392 799 &     33 590.59 s.\\ 
$\alpha\beta$ TT       &     19 163 127 770 &     43 406.79 s.\\ 
ID $\alpha\beta$ TT MC &      3 866 853 361 &     11 319.39 s.\\ 
$\alpha\beta$ TT MC    &  \bf 3 499 173 137 &  \bf 9 243.66 s.\\ 
  \end{tabular}
\end{table}

Table \ref{tableBreakthrough} gives the results for
Breakthrough. Using MC improves much the solving time. $\alpha\beta$
with MC uses seven times less nodes than the previous algorithm that 
solved Breakthrough $5 \times 5$ without patterns (i.e. parallel PN$^2$ with 64
clients \cite{Saffidine2011Solving}). Using endgame patterns divides
by seven the number of required nodes for parallel PN$^2$.

\begin{table}
  \centering
  \caption{Different algorithms for solving Misere Breakthrough.}
  \label{tableMisereBreakthrough}
  \begin{tabular}{lrrrrrrrrrr}

                       &                    &                \\ 
Size                   &       $4 \times 5$ &                \\ 
Result                 &               Lost &                \\ 
                       &              Moves &           Time \\ 
$PN^{2}$                & $>$ 42 630 000 000 &   $>$ 86 400 s.\\ 
ID $\alpha\beta$ TT    & $>$ 43 350 000 000 &   $>$ 86 400 s.\\ 
$\alpha\beta$ TT       & $>$ 42 910 000 000 &   $>$ 86 400 s.\\ 
ID $\alpha\beta$ TT MC &      1 540 153 635 &     3 661.50 s.\\ 
$\alpha\beta$ TT MC    &    \bf 447 879 697 & \bf 1 055.32 s.\\ 
  \end{tabular}
\end{table}

Table \ref{tableMisereBreakthrough} gives the results for Misere
Breakthrough. $\alpha\beta$ TT MC is the best algorithm and is much
better than all non MC algorithms.
 
\begin{table}
  \centering
  \caption{Different algorithms for solving Knightthrough.}
  \label{tableKnightthrough}
  \begin{tabular}{lrrrrrrrrrr}

                       &                    &               \\ 
Size                   &       $6 \times 6$ &               \\ 
Result                 &                Won &               \\ 
                       &              Moves &          Time \\ 
$PN^{2}$                & $>$ 33 110 000 000 &  $>$ 86 400 s.\\ 
ID $\alpha\beta$ TT    &      1 153 730 169 &    4 894.69 s.\\ 
$\alpha\beta$ TT       &      2 284 038 427 &    6 541.08 s.\\
ID $\alpha\beta$ TT MC &     \bf 17 747 503 &  \bf 102.60 s.\\ 
$\alpha\beta$ TT MC    &        528 783 129 &    1 699.01 s.\\ 
                       &                    &               \\ 
Size                   &       $7 \times 6$ &               \\ 
Result                 &                Won &               \\ 
                       &              Moves &          Time \\ 
$PN^{2}$                & $>$ 30 090 000 000 &   $>$ 86 400 s.\\ 
ID $\alpha\beta$ TT    & $>$ 17 500 000 000 &   $>$ 86 400 s.\\ 
$\alpha\beta$ TT       & $>$ 29 980 000 000 &   $>$ 86 400 s.\\ 
ID $\alpha\beta$ TT MC &  \bf 2 540 383 012 & \bf 13 716.36 s.\\ 
$\alpha\beta$ TT MC    &      6 650 804 159 &    23 958.04 s.\\ 
  \end{tabular}
\end{table}

The results for Knightthrough are in
table \ref{tableKnightthrough}. ID $\alpha\beta$ TT MC is the best 
algorithm and far better than algorithms not using MC. This is the 
first time Knightthrough $7 \times 6$ is solved.

\begin{table}
  \centering
  \caption{Different algorithms for solving Misere Knightthrough.}
  \label{tableMisereKnightthrough}
  \begin{tabular}{lrrrrrrrrrr}

                       &                    &                 \\ 
Size                   &       $5 \times 5$ &                 \\ 
Result                 &               Lost &                 \\ 
                       &              Moves &            Time \\ 
$PN^{2}$                & $>$ 45 290 000 000 &   $>$ 86 400  s.\\ 
ID $\alpha\beta$ TT    & $>$ 52 640 000 000 &    $>$ 86 400 s.\\ 
$\alpha\beta$ TT       & $>$ 56 230 000 000 &    $>$ 86 400 s.\\ 
ID $\alpha\beta$ TT MC & $>$ 41 840 000 000 &    $>$ 86 400 s.\\ 
$\alpha\beta$ TT MC    & \bf 20 375 687 163 & \bf 42 425.41 s.\\ 
  \end{tabular}
\end{table}

Table \ref{tableMisereKnightthrough} gives the results for Misere
Knightthrough. Misere Knightthrough $5 \times 5$ is solved in 20 375
687 163 moves and 42 425.41 seconds by $\alpha\beta$ TT MC. This is
the first time Misere Knightthrough $5 \times 5$ is solved. Misere
Knightthrough $5 \times 5$ is much more difficult to solve than
Knightthrough $5 \times 5$ which is solved in seconds by ID
$\alpha\beta$ TT MC. This is due to Misere Knightthrough being a 
waiting game with longer games than Knightthrough.

\begin{table}
  \centering
  \caption{Different algorithms for solving Domineering.}
  \label{tableDomineering}
  \begin{tabular}{lrrrrrrrrrr}

                       &                    &                \\ 
Size                   &       $7 \times 7$ &                \\ 
Result                 &                Won &                \\ 
                       &              Moves &           Time \\ 
$PN^{2}$                & $>$ 41 270 000 000 &   $>$ 86 400 s.\\ 
ID $\alpha\beta$ TT    &     18 958 604 687 &     35 196.62 s.\\ 
$\alpha\beta$ TT       &        197 471 137 &       376.23 s.\\ 
ID $\alpha\beta$ TT MC &      2 342 641 133 &     5 282.06 s.\\ 
$\alpha\beta$ TT MC    &     \bf 29 803 373 &   \bf 123.76 s.\\ 
  \end{tabular}
\end{table}

Table \ref{tableDomineering} gives the results for Domineering. The
best algorithm is $\alpha\beta$ TT MC which is 3 times faster than
$\alpha\beta$ TT without MC.

\begin{table}
  \centering
  \caption{Different algorithms for solving Misere Domineering.}
  \label{tableMisereDomineering}
  \begin{tabular}{lrrrrrrrrrr}

                       &                    &                \\ 
Size                   &       $7 \times 7$ &                \\ 
Result                 &                Won &                \\ 
                       &              Moves &           Time \\ 
$PN^{2}$                & $>$ 44 560 000 000 &   $>$ 86 400 s.\\ 
ID $\alpha\beta$ TT    & $>$ 49 290 000 000 &   $>$ 86 400 s.\\ 
$\alpha\beta$ TT       & $>$ 49 580 000 000 &   $>$ 86 400 s.\\ 
ID $\alpha\beta$ TT MC &      7 013 298 932 &    14 936.03 s.\\ 
$\alpha\beta$ TT MC    &     \bf 72 728 678 &   \bf 212.25 s.\\ 
  \end{tabular}
\end{table}

Table \ref{tableMisereDomineering} gives the results for Misere
Domineering. The best algorithm is $\alpha\beta$ TT MC which is much
better than all non MC algorithms.


\section{Conclusion}

For the games we solved, Misere Games are more difficult to solve than
normal games. In Misere Games the player waits and tries to force the
opponent to play a losing move. This makes the game longer and reduces
the number of winning sequences and winning moves.

Monte Carlo Move Ordering improves much the speed of $\alpha\beta$ with transposition table compare to depth first $\alpha\beta$ and Iterative Deepening $\alpha\beta$ with transposition table but without Monte Carlo Move Ordering. The experimental results show significant improvements for nine different games.

In future work we plan to parallelize the algorithms and apply them to
other problems. It would also be interesting to test if improved move ordering due to Monte Carlo Move Ordering would improve other popular solving algorithms such as DFPN. The ultimate goal with this kind of algorithms could be to solve exactly the game of Chess which is possible provided we have a very strong move ordering algorithm \cite{lemoinen}.

\bibliographystyle{named}
\bibliography{main}

\end{document}